\documentclass[11pt,a4paper]{article}
\usepackage[hyperref]{acl2017}
\usepackage{times}
\usepackage{latexsym}
\usepackage{graphicx}
\usepackage{xspace}
\usepackage[normalem]{ulem}
\usepackage{url}
\usepackage{arydshln}

\newcommand{\baseline}{Baseline\xspace}
\newcommand{\oneParaphrase}{One Paraphrase\xspace}
\newcommand{\egNone}{No Examples\xspace}
\newcommand{\egLexical}{Lexical Examples\xspace}
\newcommand{\egMixed}{Mixed Examples\xspace}
\newcommand{\novelty}{Novelty Bonus\xspace}
\newcommand{\noBonus}{No Bonus\xspace}
\newcommand{\chain}{Chain\xspace}
\newcommand{\answers}{Answers\xspace}
\newcommand{\advising}{\textsc{advising}\xspace}
\newcommand{\geoquery}{\textsc{geoquery}\xspace}
\newcommand{\wsj}{\textsc{wsj}\xspace}
\newcommand{\ubuntu}{\textsc{ubuntu}\xspace}

\newcommand{\myeg}{e.g.\@\xspace}

\aclfinalcopy 
\def\aclpaperid{3296} 

\hypersetup{
breaklinks=true,
pdftitle={Understanding Task Design Trade-offs in Crowdsourced Paraphrase Collection},
pdfauthor={Youxuan Jiang, Jonathan K. Kummerfeld and Walter S. Lasecki},
pdfsubject={Natural Language Processing},
pdfpagelayout=OneColumn}

\title{Understanding Task Design Trade-offs in\\ Crowdsourced Paraphrase Collection}

\author{Youxuan Jiang, Jonathan K. Kummerfeld \and Walter S. Lasecki \\
  Computer Science \& Engineering \\
  University of Michigan, Ann Arbor \\
  {\tt \{lyjiang,jkummerf,wlasecki\}@umich.edu} \\}

\date{}

\begin{document}
\maketitle

\begin{abstract}
  Linguistically diverse datasets are critical for training and evaluating robust machine learning systems, but data collection is a costly process that often requires experts.
  Crowdsourcing the process of paraphrase generation is an effective means of expanding natural language datasets, but there has been limited analysis of the trade-offs that arise when designing tasks.
  In this paper, we present the first systematic study of the key factors in crowdsourcing paraphrase collection.
  We consider variations in instructions, incentives, data domains, and workflows.
  We manually analyzed paraphrases for correctness, grammaticality, and linguistic diversity.
  Our observations provide new insight into the trade-offs between accuracy and diversity in crowd responses that arise as a result of task design, providing guidance for future paraphrase generation procedures.
\end{abstract}

\section{Introduction}
Paraphrases are useful for a range of tasks, including machine translation evaluation~\cite{kauchak-barzilay:2006:HLT-NAACL06-Main}, semantic parsing~\cite{wang-berant-liang:2015:ACL-IJCNLP}, and question answering~\cite{fader-zettlemoyer-etzioni:2013:ACL2013}.
Crowdsourcing has been widely used as a scalable and cost-effective means of generating paraphrases~\cite{NEGRI12,Wang-Bohus-Kamar-Horvitz:2012,tschirsich-hintz:2013:LAW7-ID}, but there has been limited analysis of the factors influencing diversity and correctness of the paraphrases workers write.

In this paper, we perform a systematic investigation of design decisions for crowdsourcing paraphrases, including the first exploration of worker incentives for paraphrasing.
For worker incentives, we either provide a bonus payment when a paraphrase is novel (encouraging diversity) or when it matches a paraphrase from another worker (encouraging agreement/correctness).
We also varied the type of example paraphrases shown to workers, the number of paraphrases requested from each worker per sentence, the subject domain of the data, whether to show answers to questions, and whether the prompt sentence is the same for multiple workers or varies, with alternative prompts drawn from the output of other workers.

Effective paraphrasing has two desired properties: correctness and diversity.
To measure correctness, we hand-labeled all paraphrases with semantic equivalence and grammaticality scores.
For diversity, we measure the fraction of paraphrases that are distinct, as well as Paraphrase In N-gram Changes~(PINC), a measure of n-gram variation.
We have released all 2,600 paraphrases along with accuracy annotations.
Our analysis shows that the most important factor is how workers are primed for a task, with the choice of examples and the prompt sentence affecting diversity and correctness significantly.

\section{Related Work}
Previous work on crowdsourced paraphrase generation fits into two categories: work on modifying the creation process or workflow, and studying the effect of prompting or priming on crowd worker output.
Beyond crowdsourced generation, other work has explored using experts or automated systems to generate paraphrases.

\subsection{Workflows for Crowd-Paraphrasing} \label{sec:related:workflows}
The most common approach to crowdsourcing paraphrase generation is to provide a sentence as a prompt and request a single paraphrase from a worker.
One frequent addition is to ask a different set of workers to evaluate whether a generated paraphrase is  correct~\cite{buzek-resnik-bederson:2010:MTURK,Burrows:2013}.
\newcite{NEGRI12} also explored an alternate workflow in which each worker writes two paraphrases, which are then given to other workers as the prompt sentence, forming a binary tree of paraphrases.
They found that paraphrases deeper in the tree were more diverse, but understanding how correctness and grammaticality vary across such a tree still remains an open question.
Near real-time crowdsourcing \cite{bigham2010vizwiz} allowed \newcite{lasecki2013conversations} to elicit variations on entire conversations by providing a setting and goal to pairs of crowd workers.
Continuous real-time crowdsourcing \cite{lasecki2011legion} allows Chorus~\newcite{lasecki2013chorus} users to hold conversations with groups of crowd workers as if the crowd was a single individual, allowing for the collection of example conversations in more realistic settings.
The only prior work regarding incentives we are aware of is by \newcite{Chklovski:2005:CPC:1088622.1088644}, who collected paraphrases in a game where the goal was to match an existing paraphrase, with extra points awarded for doing so with fewer hints.
The disadvantage of this approach was that 29\% of the collected paraphrases were duplicates.
In our experiments, duplication ranged from 1\% to 13\% in each condition.

\subsection{The Effects of Priming}
When crowd workers perform a task, they are \emph{primed} (influenced) by the examples, instructions, and context that they see.
This priming can result in systematic variations in the resulting paraphrases. \newcite{mitchell-bohus-kamar:2014:W14-50} showed that providing context, in the form of previous utterances from a dialogue, only provides benefits once four or more are included.
\newcite{kumaran-densmore-kumar:2014:Coling} provided drawings as prompts, obtaining diverse paraphrases, but without exact semantic equivalence.
When each sentence expresses a small set of slot-filler predicates, \newcite{Wang-Bohus-Kamar-Horvitz:2012} found that providing the list of predicates led to slightly faster paraphrasing than giving either a complete sentence or a short sentence for each predicate. We further expand on this work by exploring how the type of examples shown affects paraphrasing.

\subsection{Expert and Automated Generation}
Finally, there are two general lines of research on paraphrasing not focused on using crowds. The first of these is the automatic collection of paraphrases from parallel data sources, such as translations of the same text or captions for the same image~\cite{ganitkevitch2013ppdb,chen-dolan:2011:ACL-HLT2011,BOUAMOR12.555,PavlickEtAl-2015:ACL:Semantics}.
These resources are extremely large, but usually (1) do not provide the strong semantic equivalence we are interested in, and (2) focus on phrases rather than complete sentences.
The second line of work explores the creation of lattices that compactly encode hundreds of thousands of paraphrases~\cite{dreyer-marcu:2012:NAACL-HLT,Bojar2013}.
Unfortunately, these lattices are typically expensive to produce, taking experts one to three hours per sentence.

\section{Experimental Design}
We conducted a series of experiments to investigate factors in crowdsourced paraphrase creation. To do so in a controlled manner, we studied a single variation per condition.

\begin{figure}
\small
\fbox{
\parbox{0.95\linewidth}{
\begin{center}
    \textbf{Paraphrase/Reword Sentences}
\end{center}

For each sentence below, please write 2 new sentence that express the same meaning in different ways (paraphrase/reword). \\

For example: 'Which 400 level courses don't have labs?' could be rewritten as:
\begin{itemize}
  \vspace{-0.4pc}
  \item Of all the 400 level courses, which ones do not include labs?
  
  \vspace{-0.5pc}
  \item What are the 400 level courses without lab sessions?
\end{itemize}

BONUS: You will receive 5 cents bonus for each sentence you write that matches one written by another worker on the task.
\vspace{0.4pc}
}}
    \vspace{-0.4pc}
    \caption{\label{fig:instructions} Baseline task instructions.}
    \vspace{-0.6pc}
\end{figure}
    
\subsection{Definition of Valid Paraphrases}
This project was motivated by the need for strongly equivalent paraphrases in semantic parsing datasets.
We consider two sentences paraphrases if they would have equivalent interpretations when represented as a structured query, i.e., "a pair of units of text deemed to be interchangeable"~\cite{dras1999tree}. For example:

\begin{center}
\small
\parbox{0.95\linewidth}{
Prompt: \emph{Which upper-level classes are four credits?} \\
\emph{Are there any four credit upper-level classes?} \\
}
\end{center}

We considered the above two questions as paraphrases since they are both requests for a list of classes, explicit and implicit, respectively, although the second one is a polar question and the first one is not. However:

\vspace{0.4pc}
\begin{center}
\small
\parbox{0.95\linewidth}{
Prompt:\emph{Which is easier out of EECS 378 and EECS 280?} \\
\emph{Is EECS 378 easier than EECS 280?}
}
\end{center}
\vspace{0.7pc}

We did not consider the above two questions as paraphrases since the first one is requesting one of two class options and the second one is requesting a yes or no answer.

\subsection{Baseline}
We used Amazon Mechanical Turk, presenting workers with the instructions and examples in Figure~\ref{fig:instructions}.
Workers were shown prompt sentences one at a time, and asked to provide two paraphrases for each.
To avoid confusion or training effects between different conditions, we only allowed workers to participate once across all conditions.
The initial instructions shown to workers were the same across all conditions (variations were only seen after a worker accepted the task).

Workers were paid 5 cents per paraphrase they wrote plus, once all workers were done, a 5 cent bonus for paraphrases that matched another worker's paraphrase in the same condition. 
While we do not actually want duplicate paraphrases, this incentive may encourage workers to more closely follow the instructions, producing grammatical and correct sentences.
We chose this payment rate to give around minimum wage, estimating time based on prior work.

\subsection{Conditions}

\paragraph{Examples}
We provided workers with an example prompt sentence and two paraphrases, as shown in Figure~\ref{fig:instructions}.
We showed either: no examples (\egNone), two examples with lexical changes only (\egLexical), one example with lexical changes and one with syntactic changes (\egMixed), or two examples that each contained both lexical and syntactic changes (\baseline).
The variations between these conditions may prime workers differently, leading them to generate different paraphrases.

\paragraph{Incentive}
The 5 cent bonus payment per paraphrase was either not included (\noBonus), awarded for each sentence that was a duplicate at the end of the task (\baseline), or awarded for each sentence that did not match any other worker's paraphrase (\novelty).
Bonuses that depend on other workers' actions may encourage either creativity or conformity.
We did not vary the base level of payment because prior work has found that workers’ work quality is not increased by increased financial incentives due to an “anchoring” effect relative to the base rate we define \cite{mason2010financial}.

\paragraph{Workflow}
We considered three variations to workflow.
First, for each sentence, we either asked workers to provide two paraphrases (\baseline), or one (\oneParaphrase).
Asking for multiple paraphrases reduces duplication (since workers will not repeat themselves), but may result in lower diversity.
Second, since our baseline prompt sentences are questions, we ran a condition with answers shown to workers (\answers).
Third, we started all conditions with the same set of prompt sentences, but once workers had produced paraphrases, we had the option to either prompt future workers with the original prompt, or to use paraphrase from another worker.
Treating sentences as points and the act of paraphrasing as creating an edge, the space can be characterized as a graph.
We prompted workers with either the original sentences only (\baseline), or formed a chain structured graph by randomly choosing a sentence that was (1) not a duplicate, and (2) furthest from the original sentence (\chain).
These changes could impact paraphrasing because the prompt sentence is a form of priming.

\paragraph{Data domains}
We ran with five data sources: questions about university courses (\baseline), messages from dialogues between two students in a simulated academic advising session (\advising), questions about US geography~\cite[\geoquery][]{GeoQuery}, text from the Wall Street Journal section of the Penn Treebank~\cite[\wsj][]{ptb}, and discussions on the Ubuntu IRC channel (\ubuntu). We randomly selected 20 sentences as prompts from each data source with the lengths representative of the sentence length distribution in that source.

\subsection{Metrics}
\label{sec:metrics}

\paragraph{Semantic Equivalence}
For a paraphrase to be valid, its meaning must match the original sentence.
To assess this match, two of the authors---one native speaker and one non-native but fluent speaker---rated every sentence independently, then discussed every case of disagreement to determine a consensus judgement.
Prior to the consensus-finding step, the inter-annotator agreement kappa scores were .50 for correctness (moderate agreement), and .36 for grammaticality (fair agreement)~\cite{altman1990practical}.
For the results in Table~\ref{tab:results}, we used a $\chi^2$ test to measure significance, since this is a binary classification process.

\paragraph{Grammaticality}
We also judged whether the sentences were grammatical, again with two annotators rating every sentence and resolving disagreements.
Again, since this was a binary classification, we used a $\chi^2$ test for significance.

\paragraph{Time}
The time it takes to write paraphrases is important for estimating time-to-completion, and ensuring workers receive fair payment.
We measured the time between when a worker submitted one pair of paraphrases and the next.
The first paraphrase was excluded since it would skew the data by including the time spent reading the instructions and understanding the task.
We report the median time to avoid skewing due to outliers, \myeg a value of five minutes when a worker probably took a break.
We apply Mood's Median test for statistical significance.

\paragraph{Diversity}
We use two metrics for diversity, measured over correct sentences only.
First, a simple measurement of exact duplication: the number of distinct paraphrases divided by the total number of paraphrases, as a percentage (Distinct).
Second, a measure of n-gram diversity~\cite[PINC][]{chen-dolan:2011:ACL-HLT2011}\footnote{
We also considered BLEU~\cite{papineni-EtAl:2002:ACL}, which measures n-gram overlap and is used as a proxy for correctness in MT.
As expected, it strongly correlated with PINC.
}.
In both cases, a higher score means greater diversity.
For PINC, we used a t-test for statistical significance, and for Distinct we used a permutation test.

\section{Results}
We collected 2600 paraphrases: 10 paraphrases per sentence, for 20 sentences, for each of the 13 conditions.
The cost, including initial testing, was \$196.30, of which \$20.30 was for bonus payments. Table~\ref{tab:results} shows the results for all metrics.

\begin{table}[t]
    \centering
    \small
    \setlength{\tabcolsep}{3.5pt}
    \renewcommand{\arraystretch}{1.15}
    \begin{tabular}{lccccc}
        \hline
                  & \multicolumn{2}{c}{Accuracy (\%)} & Time & \multicolumn{2}{c}{Diversity} \\
        Condition & Corr & Gram & (s) & Distinct & PINC \\
        \hline
        \hline
        \baseline      &         74  &         97  &         36  &         99  &         68 \\
        \hline                                                         
        \egLexical     & \textbf{90}$\dagger$ &         98  & \textbf{27} & \textbf{93} & \textbf{55}$\dagger$ \\
        \egMixed       & \textbf{89}$\dagger$ &         96  &         36  & \textbf{87}$\dagger$ & \textbf{58}$\dagger$ \\
        \egNone        &         84  &         96  &         30  &         95  &         63  \\
        \hdashline
        \novelty       &         72  &         96  &         30  &         99  &         69  \\
        \noBonus       &         78  &         94  &         28  &         99  &         66  \\
        \hdashline
        \oneParaphrase &         82  & \textbf{89} &         38  &         96  &         65  \\
        \chain         &         68  &         94  & \textbf{25} &         98  & \textbf{74} \\
        \answers       &         80  &         94  & \textbf{29} &         96  &         65  \\
        \hdashline
        \advising      &         78  &         94  &         31  &  97 & 70 \\
        \geoquery      &         77  & \textbf{85}$\dagger$ & \textbf{25}$\dagger$ & \textbf{94} & \textbf{63} \\
        \wsj           &         68  & \textbf{90} & \textbf{61}$\dagger$ & \textbf{94}$\dagger$ & \textbf{38}$\dagger$ \\
        \ubuntu        & \textbf{56}$\dagger$ &         92  &         44  & 97 & 67 \\
        \hline
    \end{tabular}
    \vspace{-0.7pc}
    \caption{
        Variation across conditions for a range of metrics (defined in \S~\ref{sec:metrics}).
        Bold indicates a statistically significant difference compared to the baseline at the $0.05$ level, and a $\dagger$ indicates significance at the $0.01$ level, both after applying the Holm–-Bonferroni method across each row~\cite{Holm}.
    }
    \label{tab:results}
    \vspace{-0.1pc}
\end{table}

\subsection{Discussion: Task Variation}
Qualitatively, we observed a wide variety of lexical and syntactic changes, as shown by these example prompts and paraphrases (one low PINC and one high PINC in each case):

\begin{center}
\small
\parbox{0.95\linewidth}{
Prompt: \emph{How long has EECS 280 been offered for?} \\
\emph{How long has EECS 280 been offered?} \\
\emph{EECS 280 has been in the course listings how many years?}
}
\end{center}

\begin{center}
\small
\parbox{0.95\linewidth}{
Prompt: \emph{Can I take 280 on Mondays and Wednesdays?} \\
\emph{On Mondays and Wednesdays, can I take 280?} \\
\emph{Is 280 available as a Monday/Wednesday class?}
}
\end{center}
\vspace{0.2pc}

There was relatively little variation in grammaticality or time across the conditions.
The times we observed are consistent with prior work: \myeg \newcite{wang-berant-liang:2015:ACL-IJCNLP} report ${\sim}28$ sec/paraphrase.

Priming had a major impact, with the shift to lexical examples leading to a significant improvement in correctness, but much lower diversity.
The surprising increase in correctness when providing no examples has a p-value of 0.07 and probably reflects random variation in the pool of workers.
Meanwhile, changing the incentives by providing either a bonus for novelty, or no bonus at all, did not substantially impact any of the metrics.

Changing the number of paraphrases written by each worker did not significantly impact diversity (we worried that collecting more than one may lead to a decrease).
We further confirmed this by calculating PINC between the two paraphrases provided by each user, which produced scores similar to comparing with the prompt.
However, the \oneParaphrase condition did have lower grammaticality, emphasizing the value of evaluating and filtering out workers who write ungrammatical paraphrases.

Changing the source of the prompt sentence to create a chain of paraphrases led to a significant increase in diversity.
This fits our intuition that the prompt is a form of priming.
However, correctness decreases along the chain, suggesting the need to check paraphrases against the original sentence during the overall process, possibly using other workers as described in \S~\ref{sec:related:workflows}.
Meanwhile, showing the answer to the question being paraphrased did not significantly affect correctness or diversity, and in 2.5\% of cases workers incorrectly used the answer as part of their paraphrase.

We also analyzed the distribution of incorrect or ungrammatical paraphrases by worker.
7\% of workers accounted for 25\% of incorrect paraphrases, while the best 30\% of workers made no mistakes at all.
Similarly, 8\% of workers wrote 50\% of the ungrammatical paraphrases, while 70\% of workers wrote only grammatical paraphrases.
Many crowdsourcing tasks address these issues by showing workers some gold standard instances, to evaluate workers' performance during annotation.
Unfortunately, in paraphrasing there is no single correct answer, though other workers could be used to check outputs.

Finally, we checked the distribution of incorrect paraphrases per prompt sentence.
Two prompts accounted for 22\% of incorrect paraphrases:

\vspace{0.2pc}
\begin{center}
\small
\parbox{0.95\linewidth}{
Prompt:\emph{Which is easier out of EECS 378 and EECS 280?} \\
\emph{Is EECS 378 easier than EECS 280?}
}
\end{center}
\vspace{0.2pc}
\begin{center}
\small
\parbox{0.95\linewidth}{
Prompt: \emph{Is Professor Stout the only person who teaches Algorithms?} \\
\emph{Are there professors other than Stout who teach Algorithms?}
}
\end{center}
\vspace{0.3pc}

These paraphrases are not semantically equivalent to the original question, but they would elicit equivalent information, which explains why workers provided them. Providing negative examples may help guide workers to avoid such mistakes.

\subsection{Discussion: Domains}

The bottom section of Table~\ref{tab:results} shows measurements using the baseline setup, but with variations in the source domain of data.
The only significant change in correctness is on \ubuntu, which is probably due to the extensive use of jargon in the dataset, for example:

\vspace{0.2pc}
\begin{center}
\small
\parbox{0.95\linewidth}{
Prompt: \emph{ok, what does journalctl show} \\
\emph{That journalistic show is about what?}
}
\end{center}
\vspace{0.2pc}

For grammaticality, \geoquery is particularly low; common mistakes included confusion between singular/plural and has/have.
\wsj is the domain with the greatest variations.
It has considerably longer sentences on average, which explains the greater time taken.
This could also explain the lower distinctness and PINC score, because workers would often retain large parts of the sentence, sometimes re-arranged, but otherwise unchanged.

\section{Conclusion}
While previous work has used crowdsourcing to generate paraphrases, we perform the first systematic study of factors influencing the process. We find that the most substantial variations are caused by priming effects: using simpler examples leads to lower diversity, but more frequent semantic equivalence.
Meanwhile, prompting workers with paraphrases collected from other workers (rather than re-using the original prompt) increases diversity. Our findings provide clear guidance for future paraphrase generation, supporting the creation of larger, more diverse future datasets.

\vspace{1pc}
\section{Acknowledgements}
We would like to thank the members of the UMich/IBM Sapphire project, as well as all of our study participants and the anonymous reviewers for their helpful suggestions on this work.

This material is based in part upon work supported by IBM under contract 4915012629 . Any opinions, findings, conclusions or recommendations expressed above are those of the authors and do not necessarily reflect the views of IBM.

\vspace{1pc}
\bibliography{acl17paraphrase}

\begin{thebibliography}{}
\expandafter\ifx\csname natexlab\endcsname\relax\def\natexlab#1{#1}\fi

\bibitem[{Altman(1990)}]{altman1990practical}
Douglas~G Altman. 1990.
\newblock {\em Practical statistics for medical research\/}.
\newblock CRC press.
\newblock
  \href{https://www.crcpress.com/Practical-Statistics-for-Medical-Research/Altman/p/book/9780412276309}{https://www.crcpress.com/Practical-Statistics-for-Medical-Research/Altman/p/book/9780412276309}.

\bibitem[{Bigham et~al.(2010)Bigham, Jayant, Ji, Little, Miller, Miller,
  Miller, Tatarowicz, White, White et~al.}]{bigham2010vizwiz}
Jeffrey~P Bigham, Chandrika Jayant, Hanjie Ji, Greg Little, Andrew Miller,
  Robert~C Miller, Robin Miller, Aubrey Tatarowicz, Brandyn White, Samual
  White, et~al. 2010.
\newblock \href{http://dl.acm.org/citation.cfm?id=1866029.1866080}{Vizwiz:
  nearly real-time answers to visual questions}.
\newblock In {\em Proc. of the 23nd annual ACM symposium on User interface
  software and technology\/}. ACM, pages 333--342.
\newblock
  \href{http://dl.acm.org/citation.cfm?id=1866029.1866080}{http://dl.acm.org/citation.cfm?id=1866029.1866080}.

\bibitem[{Bojar et~al.(2013)Bojar, Mach{\'a}{\v{c}}ek, Tamchyna, and
  Zeman}]{Bojar2013}
Ond{\v{r}}ej Bojar, Matou{\v{s}} Mach{\'a}{\v{c}}ek, Ale{\v{s}} Tamchyna, and
  Daniel Zeman. 2013.
\newblock {\em Scratching the surface of possible translations\/}, Springer
  Berlin Heidelberg, Berlin, Heidelberg, pages 465--474.
\newblock
  \href{http://dx.doi.org/10.1007/978-3-642-40585-3\_59}{http://dx.doi.org/10.1007/978-3-642-40585-3\_59}.

\bibitem[{Bouamor et~al.(2012)Bouamor, Max, Illouz, and Vilnat}]{BOUAMOR12.555}
Houda Bouamor, Aurélien Max, Gabriel Illouz, and Anne Vilnat. 2012.
\newblock
  \href{http://www.lrec-conf.org/proceedings/lrec2012/pdf/555\_Paper.pdf}{A
  contrastive review of paraphrase acquisition techniques}.
\newblock In {\em Proc. of the Eight International Conference on Language
  Resources and Evaluation (LREC'12)\/}.
\newblock
  \href{http://www.lrec-conf.org/proceedings/lrec2012/pdf/555\_Paper.pdf}{http://www.lrec-conf.org/proceedings/lrec2012/pdf/555\_Paper.pdf}.

\bibitem[{Burrows et~al.(2013)Burrows, Potthast, and Stein}]{Burrows:2013}
Steven Burrows, Martin Potthast, and Benno Stein. 2013.
\newblock \href{http://doi.acm.org/10.1145/2483669.2483676}{Paraphrase
  acquisition via crowdsourcing and machine learning}.
\newblock {\em ACM Transactions on Intelligent Systems and Technology\/}
  4(3):43:1--43:21.
\newblock
  \href{http://doi.acm.org/10.1145/2483669.2483676}{http://doi.acm.org/10.1145/2483669.2483676}.

\bibitem[{Buzek et~al.(2010)Buzek, Resnik, and
  Bederson}]{buzek-resnik-bederson:2010:MTURK}
Olivia Buzek, Philip Resnik, and Ben Bederson. 2010.
\newblock \href{http://www.aclweb.org/anthology/W10-0735}{Error driven
  paraphrase annotation using mechanical turk}.
\newblock In {\em Proc. of the NAACL HLT 2010 Workshop on Creating Speech and
  Language Data with Amazon's Mechanical Turk\/}.
\newblock
  \href{http://www.aclweb.org/anthology/W10-0735}{http://www.aclweb.org/anthology/W10-0735}.

\bibitem[{Chen and Dolan(2011)}]{chen-dolan:2011:ACL-HLT2011}
David Chen and William Dolan. 2011.
\newblock \href{http://www.aclweb.org/anthology/P11-1020}{Collecting highly
  parallel data for paraphrase evaluation}.
\newblock In {\em Proc. of the 49th Annual Meeting of the Association for
  Computational Linguistics: Human Language Technologies\/}.
\newblock
  \href{http://www.aclweb.org/anthology/P11-1020}{http://www.aclweb.org/anthology/P11-1020}.

\bibitem[{Chklovski(2005)}]{Chklovski:2005:CPC:1088622.1088644}
Timothy Chklovski. 2005.
\newblock \href{http://doi.acm.org/10.1145/1088622.1088644}{Collecting
  paraphrase corpora from volunteer contributors}.
\newblock In {\em Proc. of the 3rd International Conference on Knowledge
  Capture\/}.
\newblock
  \href{http://doi.acm.org/10.1145/1088622.1088644}{http://doi.acm.org/10.1145/1088622.1088644}.

\bibitem[{Dras(1999)}]{dras1999tree}
Mark Dras. 1999.
\newblock {\em Tree adjoining grammar and the reluctant paraphrasing of
  text\/}.
\newblock Ph.D. thesis, Macquarie University NSW 2109 Australia.
\newblock
  \href{http://web.science.mq.edu.au/~madras/papers/thesis.pdf}{http://web.science.mq.edu.au/~madras/papers/thesis.pdf}.

\bibitem[{Dreyer and Marcu(2012)}]{dreyer-marcu:2012:NAACL-HLT}
Markus Dreyer and Daniel Marcu. 2012.
\newblock \href{http://www.aclweb.org/anthology/N12-1017}{{HyTER}:
  Meaning-equivalent semantics for translation evaluation}.
\newblock In {\em Proc. of the 2012 Conference of the North American Chapter of
  the Association for Computational Linguistics: Human Language
  Technologies\/}.
\newblock
  \href{http://www.aclweb.org/anthology/N12-1017}{http://www.aclweb.org/anthology/N12-1017}.

\bibitem[{Fader et~al.(2013)Fader, Zettlemoyer, and
  Etzioni}]{fader-zettlemoyer-etzioni:2013:ACL2013}
Anthony Fader, Luke Zettlemoyer, and Oren Etzioni. 2013.
\newblock
  \href{http://www.aclweb.org/anthology/P/P13/P13-1158.pdf}{Paraphrase-driven
  learning for open question answering}.
\newblock In {\em Proc. of the 51st Annual Meeting of the Association for
  Computational Linguistics (Volume 1: Long Papers)\/}.
\newblock
  \href{http://www.aclweb.org/anthology/P/P13/P13-1158.pdf}{http://www.aclweb.org/anthology/P/P13/P13-1158.pdf}.

\bibitem[{Ganitkevitch et~al.(2013)Ganitkevitch, {Van Durme}, and
  Callison-Burch}]{ganitkevitch2013ppdb}
Juri Ganitkevitch, Benjamin {Van Durme}, and Chris Callison-Burch. 2013.
\newblock \href{http://aclweb.org/anthology/N/N13/N13-1092.pdf}{{PPDB}: The
  paraphrase database}.
\newblock In {\em Proc. of the 2013 Conference of the North American Chapter of
  the Association for Computational Linguistics: Human Language
  Technologies\/}.
\newblock
  \href{http://aclweb.org/anthology/N/N13/N13-1092.pdf}{http://aclweb.org/anthology/N/N13/N13-1092.pdf}.

\bibitem[{Holm(1979)}]{Holm}
Sture Holm. 1979.
\newblock \href{http://www.jstor.org/stable/4615733}{A simple sequentially
  rejective multiple test procedure}.
\newblock {\em Scandinavian Journal of Statistics\/} 6(2):65--70.
\newblock
  \href{http://www.jstor.org/stable/4615733}{http://www.jstor.org/stable/4615733}.

\bibitem[{Kauchak and Barzilay(2006)}]{kauchak-barzilay:2006:HLT-NAACL06-Main}
David Kauchak and Regina Barzilay. 2006.
\newblock
  \href{http://www.aclweb.org/anthology/N/N06/N06-1058.pdf}{Paraphrasing for
  automatic evaluation}.
\newblock In {\em Proc. of the Human Language Technology Conference of the
  NAACL, Main Conference\/}.
\newblock
  \href{http://www.aclweb.org/anthology/N/N06/N06-1058.pdf}{http://www.aclweb.org/anthology/N/N06/N06-1058.pdf}.

\bibitem[{Kumaran et~al.(2014)Kumaran, Densmore, and
  Kumar}]{kumaran-densmore-kumar:2014:Coling}
A~Kumaran, Melissa Densmore, and Shaishav Kumar. 2014.
\newblock \href{http://www.aclweb.org/anthology/C14-1117}{Online gaming for
  crowd-sourcing phrase-equivalents}.
\newblock In {\em Proc. of COLING 2014, the 25th International Conference on
  Computational Linguistics: Technical Papers\/}.
\newblock
  \href{http://www.aclweb.org/anthology/C14-1117}{http://www.aclweb.org/anthology/C14-1117}.

\bibitem[{Lasecki et~al.(2013{\natexlab{a}})Lasecki, Kamar, and
  Bohus}]{lasecki2013conversations}
Walter~S. Lasecki, Ece Kamar, and Dan Bohus. 2013{\natexlab{a}}.
\newblock \href{http://www.aaai.org/ocs/index.php/HCOMP/
  HCOMP13/paper/view/7637}{Conversations in the crowd: Collecting data for
  task-oriented dialog learning}.
\newblock In {\em Scaling Speech, Language Understanding and Dialogue through
  Crowdsourcing Workshop at the First AAAI Conference on Human Computation and
  Crowdsourcing\/}.
\newblock \href{http://www.aaai.org/ocs/index.php/HCOMP/
  HCOMP13/paper/view/7637}{http://www.aaai.org/ocs/index.php/HCOMP/
  HCOMP13/paper/view/7637}.

\bibitem[{Lasecki et~al.(2011)Lasecki, Murray, White, Miller, and
  Bigham}]{lasecki2011legion}
Walter~S Lasecki, Kyle~I Murray, Samuel White, Robert~C Miller, and Jeffrey~P
  Bigham. 2011.
\newblock \href{http://dl.acm.org/citation.cfm?id=2047200}{Real-time crowd
  control of existing interfaces}.
\newblock In {\em Proc. of the 24th annual ACM symposium on User interface
  software and technology\/}. ACM, pages 23--32.
\newblock
  \href{http://dl.acm.org/citation.cfm?id=2047200}{http://dl.acm.org/citation.cfm?id=2047200}.

\bibitem[{Lasecki et~al.(2013{\natexlab{b}})Lasecki, Wesley, Nichols, Kulkarni,
  Allen, and Bigham}]{lasecki2013chorus}
Walter~S Lasecki, Rachel Wesley, Jeffrey Nichols, Anand Kulkarni, James~F
  Allen, and Jeffrey~P Bigham. 2013{\natexlab{b}}.
\newblock \href{http://dl.acm.org/citation.cfm?id=2502057}{Chorus: a
  crowd-powered conversational assistant}.
\newblock In {\em Proc. of the 26th annual ACM symposium on User interface
  software and technology\/}. ACM, pages 151--162.
\newblock
  \href{http://dl.acm.org/citation.cfm?id=2502057}{http://dl.acm.org/citation.cfm?id=2502057}.

\bibitem[{Marcus et~al.(1993)Marcus, Santorini, and Marcinkiewicz}]{ptb}
Mitchell~P. Marcus, Beatrice Santorini, and Mary~Ann Marcinkiewicz. 1993.
\newblock \href{http://aclweb.org/anthology/J93-2004}{Building a large
  annotated corpus of {English: The Penn Treebank}}.
\newblock {\em Computational Linguistics\/} 19(2):313--330.
\newblock
  \href{http://aclweb.org/anthology/J93-2004}{http://aclweb.org/anthology/J93-2004}.

\bibitem[{Mason and Watts(2010)}]{mason2010financial}
Winter Mason and Duncan~J Watts. 2010.
\newblock \href{http://dl.acm.org/citation.cfm?id=1600175}{Financial incentives
  and the performance of crowds}.
\newblock {\em ACM SigKDD Explorations Newsletter\/} 11(2):100--108.
\newblock
  \href{http://dl.acm.org/citation.cfm?id=1600175}{http://dl.acm.org/citation.cfm?id=1600175}.

\bibitem[{Mitchell et~al.(2014)Mitchell, Bohus, and
  Kamar}]{mitchell-bohus-kamar:2014:W14-50}
Margaret Mitchell, Dan Bohus, and Ece Kamar. 2014.
\newblock \href{http://www.aclweb.org/anthology/W14-5003}{Crowdsourcing
  language generation templates for dialogue systems}.
\newblock In {\em Proc. of the INLG and SIGDIAL 2014 Joint Session\/}.
\newblock
  \href{http://www.aclweb.org/anthology/W14-5003}{http://www.aclweb.org/anthology/W14-5003}.

\bibitem[{Negri et~al.(2012)Negri, Mehdad, Marchetti, Giampiccolo, and
  Bentivogli}]{NEGRI12}
Matteo Negri, Yashar Mehdad, Alessandro Marchetti, Danilo Giampiccolo, and
  Luisa Bentivogli. 2012.
\newblock
  \href{http://www.lrec-conf.org/proceedings/lrec2012/pdf/772\_Paper.pdf}{Chinese
  whispers: Cooperative paraphrase acquisition}.
\newblock In {\em Proc. of the Eighth International Conference on Language
  Resources and Evaluation (LREC'12)\/}.
\newblock
  \href{http://www.lrec-conf.org/proceedings/lrec2012/pdf/772\_Paper.pdf}{http://www.lrec-conf.org/proceedings/lrec2012/pdf/772\_Paper.pdf}.

\bibitem[{Papineni et~al.(2002)Papineni, Roukos, Ward, and
  Zhu}]{papineni-EtAl:2002:ACL}
Kishore Papineni, Salim Roukos, Todd Ward, and Wei-Jing Zhu. 2002.
\newblock \href{http://www.aclweb.org/anthology/P/P02/P02-1040.pdf}{{BLEU}: a
  method for automatic evaluation of machine translation}.
\newblock In {\em Proc. of the 40th Annual Meeting of the Association for
  Computational Linguistics\/}.
\newblock
  \href{http://www.aclweb.org/anthology/P/P02/P02-1040.pdf}{http://www.aclweb.org/anthology/P/P02/P02-1040.pdf}.

\bibitem[{Pavlick et~al.(2015)Pavlick, Rastogi, Ganitkevich, and Ben
  Van~Durme}]{PavlickEtAl-2015:ACL:Semantics}
Ellie Pavlick, Pushpendre Rastogi, Juri Ganitkevich, and Chris Callison-Burch
  Ben Van~Durme. 2015.
\newblock \href{http://aclweb.org/anthology/P/P15/P15-2070.pdf}{{PPDB} 2.0:
  Better paraphrase ranking, fine-grained entailment relations, word
  embeddings, and style classification}.
\newblock In {\em Proc. of the 53rd Annual Meeting of the Association for
  Computational Linguistics (ACL 2015)\/}.
\newblock
  \href{http://aclweb.org/anthology/P/P15/P15-2070.pdf}{http://aclweb.org/anthology/P/P15/P15-2070.pdf}.

\bibitem[{Tang and Mooney(2001)}]{GeoQuery}
Lappoon~R. Tang and Raymond~J. Mooney. 2001.
\newblock
  \href{https://link.springer.com/chapter/10.1007/3-540-44795-4\_40}{Using
  multiple clause constructors in inductive logic programming for semantic
  parsing}.
\newblock In {\em Proc. of the 12th European Conference on Machine Learning\/}.
\newblock
  \href{https://link.springer.com/chapter/10.1007/3-540-44795-4\_40}{https://link.springer.com/chapter/10.1007/3-540-44795-4\_40}.

\bibitem[{Tschirsich and Hintz(2013)}]{tschirsich-hintz:2013:LAW7-ID}
Martin Tschirsich and Gerold Hintz. 2013.
\newblock \href{http://www.aclweb.org/anthology/W13-2325}{Leveraging
  crowdsourcing for paraphrase recognition}.
\newblock In {\em Proc. of the 7th Linguistic Annotation Workshop and
  Interoperability with Discourse\/}.
\newblock
  \href{http://www.aclweb.org/anthology/W13-2325}{http://www.aclweb.org/anthology/W13-2325}.

\bibitem[{Wang et~al.(2012)Wang, Bohus, Kamar, and
  Horvitz}]{Wang-Bohus-Kamar-Horvitz:2012}
W.~Y. Wang, D.~Bohus, E.~Kamar, and E.~Horvitz. 2012.
\newblock \href{http://ieeexplore.ieee.org/document/6424200/}{Crowdsourcing the
  acquisition of natural language corpora: Methods and observations}.
\newblock In {\em 2012 IEEE Spoken Language Technology Workshop (SLT)\/}.
\newblock
  \href{http://ieeexplore.ieee.org/document/6424200/}{http://ieeexplore.ieee.org/document/6424200/}.

\bibitem[{Wang et~al.(2015)Wang, Berant, and
  Liang}]{wang-berant-liang:2015:ACL-IJCNLP}
Yushi Wang, Jonathan Berant, and Percy Liang. 2015.
\newblock \href{http://www.aclweb.org/anthology/P15-1129}{Building a semantic
  parser overnight}.
\newblock In {\em Proc. of the 53rd Annual Meeting of the Association for
  Computational Linguistics and the 7th International Joint Conference on
  Natural Language Processing (Volume 1: Long Papers)\/}.
\newblock
  \href{http://www.aclweb.org/anthology/P15-1129}{http://www.aclweb.org/anthology/P15-1129}.

\end{thebibliography}
\bibliographystyle{acl_natbib}

\end{document}